\documentclass{article} 
\usepackage[final]{colm2026_conference}

\usepackage{microtype}
\usepackage{hyperref}
\usepackage{url}
\usepackage{booktabs}

\usepackage{lineno}
\usepackage{graphicx}
\usepackage{amssymb}
\usepackage{subcaption}
\usepackage{wrapfig}
\usepackage{amsmath} 

\definecolor{darkblue}{rgb}{0, 0, 0.5}
\hypersetup{colorlinks=true, citecolor=darkblue, linkcolor=darkblue, urlcolor=darkblue}

\title{Let's Scale Step by Step: Compute-Efficient \\Hyperparameter Transfer for Large-Scale Mixture-of-Experts}

\renewcommand{\thefootnote}{\fnsymbol{footnote}}

\author{%
    Nayeon Kim$^1$\thanks{Contact: \texttt{\{lana.ny, mat.mul\}@kakaocorp.com}}, 
    Hojin Lee$^2$\thanks{Work done at Kakao Corp.}, 
    Yunju Bak$^1$, Jaesun Park$^1$, 
    Boseop Kim$^1$\footnotemark[1] \\
    $^1$Kakao Corp., $^2$Upstage AI
}

\begin{document}

\ifcolmsubmission
\linenumbers
\fi

\maketitle

\renewcommand{\thefootnote}{\arabic{footnote}}
\setcounter{footnote}{0}

\begin{abstract}
Mixture-of-Experts (MoE) architectures significantly expand model capacity without a proportional increase in computational cost. 
However, optimizing their hyperparameters—particularly the learning rate—at extreme scales of both model size and token budget via sweeping remains computationally prohibitive.
In this paper, we propose a compute-efficient, two-step hyperparameter transfer framework that estimates optimal learning rates for training large MoE models by transferring them across scaling model widths, and subsequently extrapolating to trillion-token horizons.
First, we formulate a Maximal Update Parameterization ($\mu$P) adaptation for MoE architectures utilizing Multi-head Latent Attention (MLA) and the Muon optimizer, demonstrating that optimal learning rates transfer consistently across width-scaled models. 
Second, we extend this transferability along the token dimension by establishing a predictive scaling law. By applying linear regression to the optimal values derived from small proxy models on limited budgets, we successfully extrapolate the ideal learning rate to massive training horizons (e.g., 10 trillion tokens) with high fidelity ($R^2=0.95$). 
Consequently, this indicates that proxy training on small models is sufficient to determine the optimal learning rate for the extensive training of large-scale MoEs.
We apply the proposed methodology to pretrain our foundation model (155B total, 17B active parameters) from scratch, and the stable training and evaluation results validate that optimal configurations for full-scale target models can be accurately predicted with minimal ablation costs.
\end{abstract}

\section{Introduction}
Mixture-of-Experts (MoE) architectures \citep{shazeer2017outrageously} have emerged as a powerful paradigm for scaling LLM capacity without a proportional increase in computational cost, enabling both greater model specialization and cost-efficient inference. 
As a result, MoE models can achieve significantly higher effective capacity than dense counterparts trained on the same data, delivering stronger performance without any additional training cost \citep{shazeer2017outrageously,artetxe2021efficient,fedus2022switch,ludziejewski2024scaling}. 
Prominent open-source LLMs such as DeepSeek-V3 \citep{liu2024deepseek}, Qwen3-235B-A22B \citep{yang2025qwen3}, Kimi-K2.5 \citep{team2026kimi}, and GLM-5 \citep{zeng2026glm} employ MoE designs to deliver powerful performance while reducing the per-token active parameters and active inference FLOPs, often outperforming dense counterparts of similar or larger size on a range of tasks.
However, MoE architectures introduce additional hyperparameters to manage expert routing and load balancing, which expands the hyperparameter space and makes tuning considerably more complex and expensive than for dense models.
Among these hyperparameters, the learning rate stands out as a critical determinant of successful training \citep{bengio2012practical,smith2018disciplined,kaplan2020scaling,yang2022tensor,li2025predictable}; yet its optimal value is highly sensitive to both model size and token budget \citep{li2025predictable,bjorck2025scaling}.
This means that every change in scale---whether in model size or training horizon---demands a fresh and expensive hyperparameter search.

\begin{figure}[t]
    \centering
    \begin{subfigure}[b]{0.49\textwidth}
        \centering
        \includegraphics[width=\textwidth]{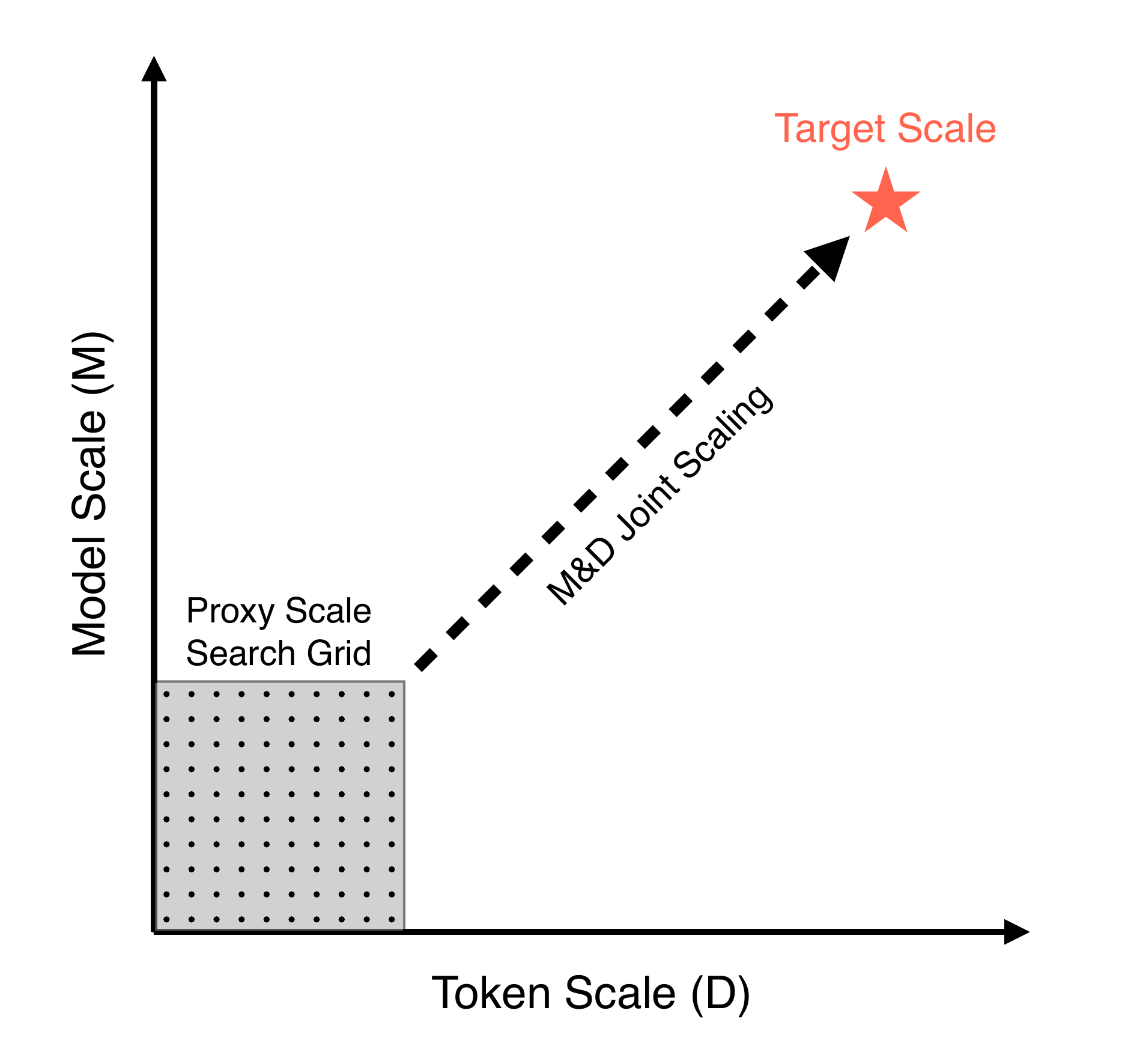} 
        \caption{Conventional Hyperparameter Scaling}
        \label{fig:main-a}
    \end{subfigure}
    \begin{subfigure}[b]{0.49\textwidth}
        \centering
        \includegraphics[width=\textwidth]{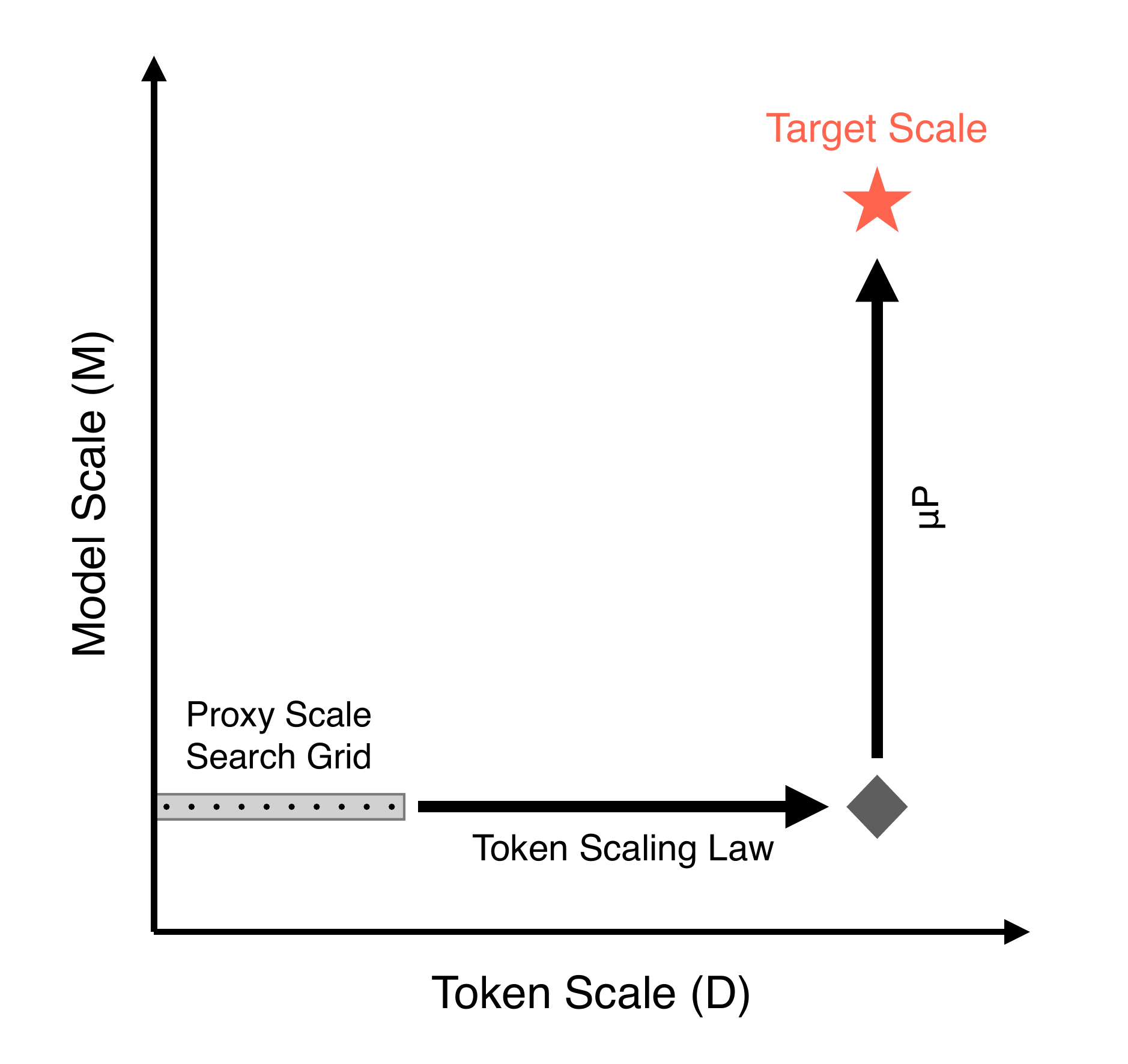} 
        \caption{Two-Step Hyperparameter Transfer}
        \label{fig:main-b}
    \end{subfigure}
    
    \caption{(a) In conventional hyperparameter optimization, exhaustive 2D sweeps across both model scale ($M$) and token scale ($D$) are required to jointly predict the target scale ($\bigstar$), incurring prohibitive computational costs. (b) Our approach decouples the two dimensions:  $\mu$P-based width transferability eliminates model-scale sweeps, while a token scaling law from a few proxy runs enables direct extrapolation to the target scale ($\bigstar$), replacing 2D sweeps with a lightweight 1D search along the token dimension.}
    \label{fig:main}
    \vspace{-15pt}    
\end{figure}

\begin{wrapfigure}{r}{0.5\textwidth}
    \vspace{-5pt}    
    \centering
    \begin{subfigure}[b]{0.23\textwidth}
        \centering
        \includegraphics[width=\textwidth]{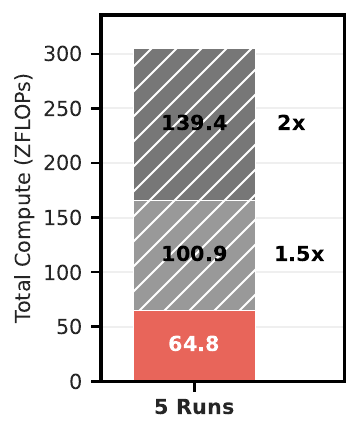} 
        \caption{1D vs. 2D Sweeps}
        \label{fig:main-2-a}
    \end{subfigure}%
    \begin{subfigure}[b]{0.235\textwidth}
        \centering
        \includegraphics[width=\textwidth]{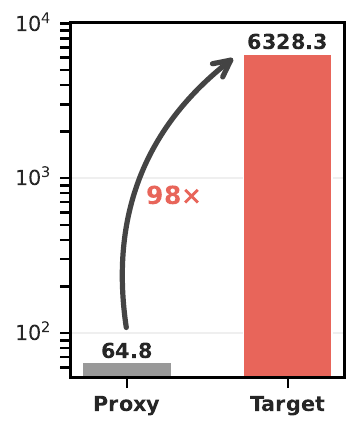} 
        \caption{Proxy vs. Target}
        \label{fig:main-2-b}
    \end{subfigure}
    \caption{\textbf{Proxy} denotes the 5 proxy-scale runs from Section \ref{lrmup10T} and \textbf{Target} denotes our large-scale training from Section \ref{ourmodel}. (a) 2D sweeps require scaling across model size, resulting in additional computation. Model scale search ($1.5\times, 2\times$ wider than \textbf{Proxy}; details in Appendix \ref{Appendix:model-configs}) incurs an additional 240.3 ZFLOPs of compute beyond the \textbf{Proxy} run cost of 64.8 ZFLOPs. (b) Total computation of the \textbf{Target} is approximately $98\times$ larger than the sum of the \textbf{Proxy} runs. FLOPs are computed following \citet{narayanan2021efficient}.}
    \label{fig:main-2}
    \vspace{-10pt}        
\end{wrapfigure}

To mitigate the prohibitive computational cost of hyperparameter sweeps at massive scales, frameworks such as Maximal Update Parameterization ($\mu$P) \citep{pmlr-v139-yang21c} and $\mu$-Transfer \citep{yang2022tensor} have been proposed. 
These methods enable the zero-shot transfer of optimal hyperparameters (e.g., learning rates) from small proxy models to larger variants.
However, these methods were originally designed for dense models, where scaling width is the primary axis. 
MoE architectures introduce an additional scaling dimension---sparsity---governed by the ratio of active to total experts. 
For MoE models at extreme parameter scales (e.g., over 100B total parameters), simply expanding the width (e.g., hidden size) becomes increasingly intractable due to prohibitive inference costs and hardware constraints. 
Instead, increasing the total number of experts while keeping active experts fixed offers a more practical path, as it expands model capacity without a proportional increase in inference compute.
Nevertheless, sparsity follows distinct scaling behaviors \citep{clark2022unified,tian2025towards,team2025kimi}, and it remains unclear whether existing hyperparameter transfer frameworks derived for width scaling generalize to the sparsity dimension. 
This critical gap remains largely unaddressed, leaving optimal hyperparameter configuration prediction for large-scale MoE training an open problem.

To bridge this gap, we propose a practical, resource-efficient two-step hyperparameter transfer framework for large-scale MoE pretraining. 
Figure \ref{fig:main} illustrates the difference between conventional hyperparameter scaling approaches and our proposed method.
Conventional approaches consider model and token scales simultaneously, thereby relying on exhaustive 2D sweeps across both dimensions to empirically predict optimal configurations at large scale.
In contrast, our approach leverages $\mu$P to enable robust transfer across model scales. 
This requires only a few small-scale proxy runs to extrapolate the optimal learning rate along the token dimension, allowing for a direct transfer to full-scale model training.
Figure \ref{fig:main-2-a} illustrates how these 2D sweeps incur computation overhead as the model scale search space grows.
Our findings suggest that it is entirely feasible to estimate optimal training hyperparameters (i.e., learning rates) for extreme scales---spanning both massive model widths and long token horizons---without resorting to costly trial-and-error sweeps. We validate our methodology by pretraining our foundation MoE model (155B total, 17B active parameters) over 10 trillion tokens from scratch. Details are summarized in Section \ref{ourmodel}.

Our main contributions are as follows:
\begin{itemize}
    \item \textbf{$\mu$P Adaptation for MoE Architectures:} Building on prior work discussing the adaptation of $\mu$P to MoE architectures, we study zero-shot hyperparameter transfer when scaling both model width and the total number of experts (i.e., sparsity expansion). We further explore the use of the Muon optimizer in this setting, extending the empirical scope of $\mu$P-based MoE scaling.
    
    \item \textbf{Token-Scale Extrapolation within a Two-Step Framework:} We introduce a two-step predictive framework to extrapolate optimal learning rates to unseen, long token horizons based on cost-effective small-scale proxy runs, significantly reducing the computational overhead of hyperparameter tuning.

    \item \textbf{Large-Scale Validation:} We apply the proposed two-step framework to pretrain a large-scale MoE foundation model (155B total, 17B active parameters) from scratch over 10 trillion tokens, validating its practical effectiveness at scale.
\end{itemize}


\section{Methods}\label{sec:2}
Our two-step hyperparameter transfer framework requires two key components: (1) formulating $\mu$P specifically for MoE architectures, and (2) establishing a scaling law that predicts the optimal learning rate as the token budget scales. 
Section \ref{sec:2.1} describes our adaptation of $\mu$P to MoE architectures, and the results in Section \ref{lrmup} validate that this formulation ensures that the optimal learning rate can be reliably transferred across model widths.
Section \ref{sec:2.2} then addresses extrapolation to extended training horizons: Section \ref{sec:2.2.1} defines the optimal learning rate at a given token scale, and Section \ref{sec:2.2.2} presents our methodology for extending these estimates to unseen, large-scale token horizons. The efficacy of this extrapolation is demonstrated in Section \ref{sec:4.3}.

\subsection{Applying $\mu$-Parameterization to MoE}\label{sec:2.1}
Although determining the optimal learning rate is critical for successful model training, conducting comprehensive hyperparameter search for large-scale MoE models exceeding 100B parameters is computationally prohibitive.
Therefore, we adopt Maximal Update Parameterization ($\mu$P) \citep{pmlr-v139-yang21c} that offers a potential solution by enabling zero-shot hyperparameter transfer. 
Following $\mu$-Transfer \citep{yang2022tensor}, Table \ref{tab:params} categorizes model parameters based on their expansion behavior under width scaling. 
Since none of our parameters have shapes invariant to model width (i.e., tied to fixed constants such as vocabulary size or context length), we exclude scalar-like parameters. 
Vector-like parameters receive $\mu$P initialization only, while matrix-like parameters are subject to both $\mu$P initialization and learning rate scaling, as detailed in Table \ref{tab:dense-mup}.
\begin{table}[t]
    \centering
    \resizebox{\textwidth}{!}{
    \begin{tabular}{p{2.5cm}p{7.2cm}p{3.3cm}}
    \toprule
    \textbf{Parameter Type} & \textbf{Definition} & \textbf{Examples} \\ \midrule
    \textbf{Vector-like} & Parameters that contain exactly one infinitely expandable dimension. & Embeddings, biases, expert FC2 weights \\
    \addlinespace
    \textbf{Matrix-like} & Parameters that contain two infinitely expandable dimensions. & FFN, attention, router, expert FC1 weights \\ \bottomrule
    \end{tabular}
    }
    \caption{Parameter classification by shape invariance to model width. For MoE layers, router parameters and expert FC1 weights are classified as matrix-like parameters, while expert FC2 weights are treated as vector-like since their effective input dimensionality remains bounded by the fixed number of active experts and the fixed MoE intermediate dimension.}
    \label{tab:params}
    \vspace{-5pt}            
\end{table}

\begin{table}[t]
    \centering
    \resizebox{\textwidth}{!}{    
    \begin{tabular}{lccc}
    \toprule
    & \textbf{I/O Embeddings \& All Biases} & \textbf{Hidden Weights} & \textbf{Hidden Weights} \\
    & \textbf{(Vector-like)} & \textbf{(Vector-like)} & \textbf{(Matrix-like)} \\
    \midrule
    \textbf{Init. Var} & 0.04 & $\mathrm{fan\_in_{\text{base}}/\mathrm{fan\_in}}$ & $\mathrm{fan\_in_{\text{base}}/\mathrm{fan\_in}}$ \\ 
    \addlinespace  
    \textbf{LR Scaling Factor} & 1 & 1 & $\mathrm{fan\_in_{\text{base}}/\mathrm{fan\_in}}$ \\ 
    \bottomrule
    \end{tabular}
    }
    \caption{$\mu$P and learning rate scaling factors for dense and MoE models. $\mathrm{fan\_in}$ and $\mathrm{fan\_in_{\text{base}}}$ are the fan-in values of the width-scaled model and the base proxy model, respectively. Learning rate scaling factors are multiplied by the learning rate when updating the corresponding parameters. Notably, for MLA, the low-rank projection dimensions for query and key-value are kept fixed during width scaling. Since these serve as the $\mathrm{fan\_in}$ of the corresponding up-projection matrices, the learning rate scaling applied to these matrices has no effect, as the scaling factor reduces to 1.}
    \label{tab:dense-mup}
    \vspace{-10pt}        
\end{table}

Following the findings of \citet{wortsman2024small}, which show that the properties of $\mu$P can be maintained by solely scaling the learning rates for linear layers, we apply learning rate scaling exclusively to matrix-like hidden weights.
As depth scaling is known to be highly unstable for $\mu$P-based hyperparameter transfer \citep{yang2022tensor, wortsman2024small}, we consider only width scaling while keeping the depth (i.e., number of layers) fixed. For attention, the head dimension remains fixed while the number of heads scales proportionally with the hidden dimension \citep{yang2022tensor,wortsman2024small}.

When scaling up the MoE models, we fix the number of active experts per token and the MoE intermediate dimension, while increasing the total number of experts and the hidden dimension. 
This reflects a practical MoE scaling strategy: rather than expanding width alone, increasing sparsity raises model capacity while keeping per-token inference costs minimal \citep{team2025kimi}. 
Consequently, sparsity becomes a key scaling axis alongside width for $\mu$P-based hyperparameter transfer.
Under this scaling path, total scale, active scale, and sparsity are coupled rather than treated as independent axes. 
This coupling is consistent with the spectral-condition view of $\mu$P: changing the active ratio by increasing the number of experts does not introduce any additional change to the fan-in or fan-out of each individual expert beyond that induced by width scaling, and therefore remains compatible with the same $\mu$P scaling rule \citep{yang2023spectral}.
This setting is further motivated by practical efficiency considerations. 
Recent MoE scaling strategies increasingly rely on lower active ratios to expand model capacity efficiently, but very high sparsity can be hardware-inefficient at proxy scale due to low arithmetic intensity. 
Running the proxy search at lower sparsity and transferring to a higher-sparsity target therefore keeps the search itself efficient while still targeting the desired sparse regime.

\subsection{Extrapolating Optimal Learning Rates to Long Token Horizons}\label{sec:2.2}
Although $\mu$P-based hyperparameter transfer significantly reduces search costs, performing a direct search for the optimal learning rate over trillions of training tokens (e.g., 10T tokens) remains computationally prohibitive even when using small proxy models.

To address this challenge, we aim to identify the optimal learning rate within a constrained token budget and extrapolate the results to the full target token horizon.
Building upon the methodology in Section \ref{sec:2.1}, the results presented in Section \ref{lrmup} verify that optimal learning rates can be successfully transferred when scaling MoE model width. 
Therefore, if a scaling rule along the token dimension is established, hyperparameter transfer from small proxy models trained on limited token budgets to large-scale models with extensive token budgets can be optimized through a two-step framework.

Importantly, our framework deliberately focuses on learning rate transfer, excluding batch size. 
This is not merely a simplification---batch size occupies a fundamentally different role: it is not only a modeling hyperparameter but also a system-level variable frequently adjusted during training to maximize hardware throughput. 
Moreover, the literature remains divided on how the optimal batch size scales, with conflicting dependencies on training compute \citep{bi2024deepseek,tian2025towards} or the token budget alone \citep{li2025predictable}. 
Similar contradictions exist in critical batch size estimations, where assumptions alternate between dependencies solely on the token budget \citep{zhang2024does} and training loss-driven dynamics \citep{li2025minimax}.
Given these contradictions and practical constraints, we fix the batch size to maximize GPU efficiency and treat the learning rate as the primary transfer target.
By decoupling learning rate optimization from batch size, we propose a learning rate scaling law that remains robust regardless of the batch size chosen to maximize throughput in a given hardware environment.

\subsubsection{Estimating Optimal Learning Rates Over Short Token Budgets}\label{sec:2.2.1}
To estimate the optimal learning rate for large-scale training, we train small proxy models over short token horizons across a range of learning rates.

We employ the Warmup-Stable-Decay (WSD) scheduler \citep{hu2024minicpm} as the target long-horizon training setup, matching the number of warmup steps. 
However, introducing learning rate decay phase for proxy experiments presents two major challenges. 
First, it is computationally expensive; even if early decay is applied at an intermediate point, each run yields only a single valid data point at its final checkpoint. Consequently, mapping out optimal learning rates across various token budgets would necessitate a large number of independent experiments. 
Second, performing premature decay during proxy training risks introducing bias into the loss estimates \citep{li2025predictable}. 
To overcome these limitations, we instead terminate proxy runs during the stable phase without decay and apply Exponential Moving Average (EMA) to the model weights. 
This approach enables us to extract multiple checkpoints at desired token intervals from a single training run, efficiently generating numerous data points representative of a wide range of token budgets.
EMA effectively smooths noise in the parameter trajectory and slows parameter updates, thereby resembling the effect of learning rate decay \citep{morales-brotons2024exponential,ernie2025technicalreport,tian2025wsm,li2026model}. 
\citet{zhang2024does} demonstrate that constant learning rate with exponential weight averaging (equivalent to EMA in our setting) performs comparably to cosine decay training in large-batch regimes, with the gap widening as the batch size decreases. 
Our training uses a large global batch size of 32M tokens, making these large-batch findings particularly relevant to our setting.
In practice, prior large-scale training works adopt closely related strategies: \citet{liu2024deepseek} relies on EMA to approximate post-decay model quality, while \citet{olmo2025olmo} uses checkpoint weight averaging to estimate model performance during pretraining without repeated learning rate annealing.

Let $\theta^{(t)}$ denote the model parameters at step $t$. The EMA parameters $\theta_{\text{EMA}}^{(t)}$ are updated as follows:

\begin{equation}\label{eq:ema}
    \theta_{\text{EMA}}^{(t)} = \alpha \cdot \theta_{\text{EMA}}^{(t-1)} + (1 - \alpha) \cdot \theta^{(t)}
\end{equation}

We set the smoothing factor to $\alpha=0.6$, placing substantial weight on recent updates while mitigating the influence of unstable dynamics during the early stages of training.
Model checkpoints are updated via EMA at approximately 2B-token intervals, and the resulting checkpoints at every 10B tokens are used for subsequent analysis. 
Based on the definition of the effective decay window by \citet{ernie2025technicalreport}, our EMA configuration implies that the most recent 20B tokens maintain more than 1\% influence on the merged weights.
To identify the optimal learning rate ($\eta^*$) for a given token budget, we evaluate the validation loss on a fixed batch across all tested learning rates using these merged checkpoints.
Following \citet{bjorck2025scaling}, for each token scale ($B$), we model the relationship between the validation loss ($\mathcal{L}$) and the log-transformed learning rate ($\log \eta$) using a second-order polynomial:

\begin{equation}
\label{eq:parabola}
\mathcal{L} (\eta) = a (\log \eta)^2 + b (\log \eta) + c
\end{equation}

The vertex of this parabola, $\log \eta^* = -b / 2a$, yields the estimated optimal learning rate ($\eta^* = \exp(-b / 2a)$) for the given token budget ($B$).

\subsubsection{Extrapolating Optimal Learning Rates Across Long Token Horizons}\label{sec:2.2.2}
We leverage the optimal learning rates estimated from short-token budget experiments via Equation \eqref{eq:parabola} to extrapolate the optimal learning rate for the full target token horizon.
Specifically, we perform a linear regression in log-log space between the estimated optimal learning rate ($\eta^*$) and the corresponding token budget ($B$):

\begin{equation}\label{eq:extrapolate}
\log(\eta^*) = \beta \cdot \log(B) + \gamma
\end{equation}

where $\beta$ and $\gamma$ represent the regression coefficients.
This regression model enables the extrapolation of the optimal learning rate to significantly larger token horizons (e.g., 10T tokens) without additional computational cost. 

In experiments that employ batch size scheduling---increasing the batch size after a predetermined number of steps---we restrict the regression to data points collected after the training dynamics have stabilized following the batch size increase. This ensures that the optimization regime is consistent across the fitted points and avoids artifacts arising from transient instabilities during the batch size transition.

\begin{figure}[!t]
    \centering
    \begin{subfigure}[b]{0.49\textwidth}
        \centering
        \includegraphics[width=\textwidth]{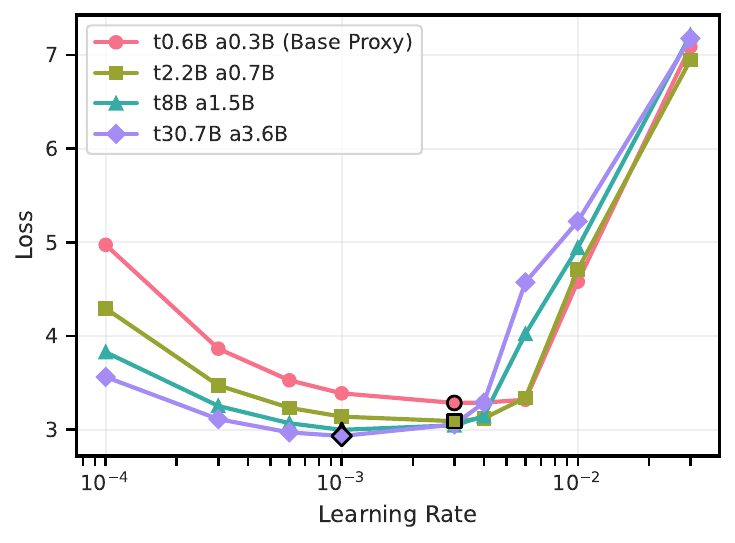} 
        \caption{SP (Standard Parameterization)}
        \label{fig:moemup-sp}
    \end{subfigure}
    \begin{subfigure}[b]{0.49\textwidth}
        \centering
        \includegraphics[width=\textwidth]{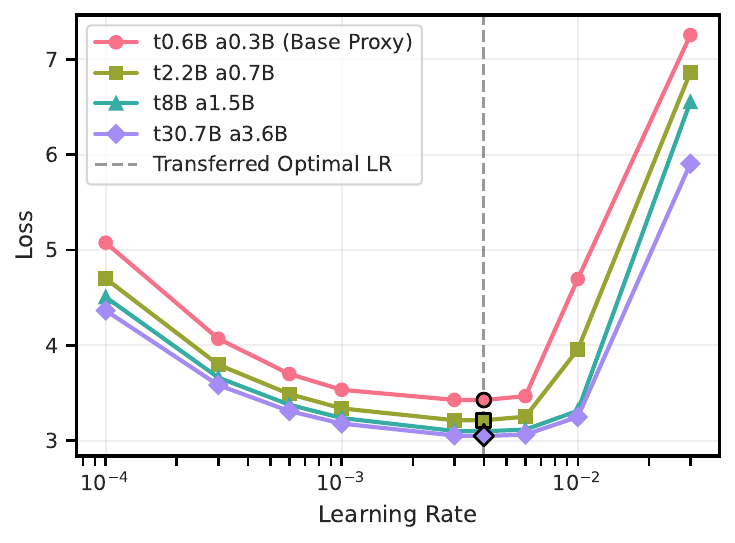} 
        \caption{$\mu$P (Maximal Update Parameterization)}
        \label{fig:moemup-mup}
    \end{subfigure}
    
    \caption{Comparison of learning rate transferability under Standard Parameterization (SP) and Maximal Update Parameterization ($\mu$P) across MLA MoE models of increasing width. Markers with black outlines indicate the learning rate achieving the lowest training loss (i.e., optimal learning rate) for each model. (a) Under SP, the optimal learning rate shifts as model width scales, failing to transfer from the base proxy model (0.6B total, 0.3B active) to larger variants. (b) Under $\mu$P, the optimal learning rate identified in the base proxy consistently transfers across models scaled to $2\times$ (2.2B total, 0.7B active), $4\times$ (8B total, 1.5B active), and $8\times$ (30.7B total, 3.6B active) the base width.}
    \label{fig:moemup}
    \vspace{-10pt}            
\end{figure}

\section{Experiments and Analysis}
\subsection{Experimental Setup}
We use the Muon optimizer \citep{jordan2024muon}, which has been shown to accelerate convergence while maintaining competitive performance compared to conventional optimizers (e.g., AdamW \citep{loshchilov2017decoupled}). 
The extended stable phase of the WSD scheduler \citep{hu2024minicpm} allows batch size scheduling to be applied effectively without disrupting training dynamics. Accordingly, batch size scheduling is applied in experiments using the WSD scheduler.
Further details of the experimental setup are provided in Appendix \ref{Appendix:training-details}. All experiments are conducted on NVIDIA H200 GPUs, using an internal fork of Megatron-LM \citep{shoeybi2019megatron}.

\subsection{$\mu$P and Learning Rate Transfer with MoE Models}\label{lrmup}
We set the base proxy MoE model's hidden dimension ($d_\text{model}$) to 256, the total number of experts ($n_\text{experts}$) to 16, and the number of attention heads to 4. 
Following state-of-the-art large MoE models such as DeepSeek-V3 \citep{liu2024deepseek} and Kimi-K2.5 \citep{team2026kimi}, we adopt Multi-head Latent Attention (MLA) \citep{liu2024deepseekv2} for all models. MLA compresses the key-value cache into a low-dimensional latent space, significantly reducing KV cache memory overhead during inference and long-context extension.
We then scale the model width by factors of $2\times$, $4\times$, and $8\times$, proportionally increasing the hidden dimension, total number of experts, and number of attention heads. 
As the total number of experts increases, the MoE routed scaling factor \citep{liu2024deepseek} is adjusted for each model size, following the gate scaling factor calculation defined by \citet{liu2025muon}. Detailed model configurations are listed in Table \ref{tab:model-configs} in Appendix \ref{Appendix:model-configs}.
All models are trained for 1.3B tokens from a general knowledge English corpus.
For each model size, we sweep the maximum learning rate over $\{1, 3, 6\} \times 10^{-4}, \{1, 3, 4, 6\} \times 10^{-3},$ and $\{1, 3\} \times 10^{-2}$ to validate transferability. 

Figure \ref{fig:moemup} demonstrates that width-scaled MoE models can successfully share optimal learning rates---defined as the learning rate yielding the lowest training loss---under $\mu$P \citep{pmlr-v139-yang21c}. In contrast, without the $\mu$P learning rate described in Section \ref{sec:2.1} (referred to as Standard Parameterization, SP), the optimal learning rate fails to transfer across model widths.
Corresponding results for dense models are included in Appendix \ref{Appendix:LR-Dense}.

\subsection{Finding Optimal Learning Rates Across Long Token Horizons}\label{lrtokenhorizon}\label{sec:4.3}

\begin{wrapfigure}{r}{0.5\textwidth}
    \vspace{-15pt}
    \centering
    \includegraphics[width=0.99\linewidth]{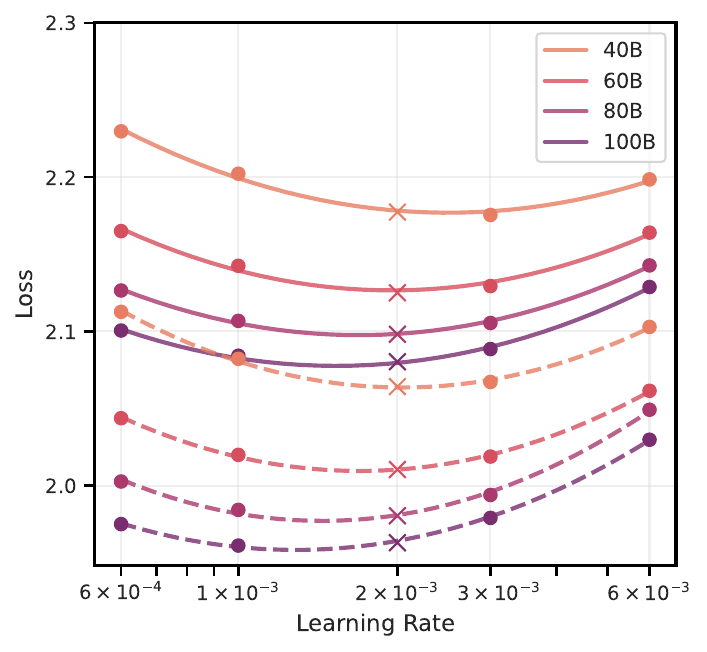}
    \caption{Solid lines ($-$) denote the base proxy model (5.6B total, 1.8B active), while dashed lines ($--$) represent the $2\times$ width-scaled held-out model (20.7B total, 3.8B active), each trained on 40B, 60B, 80B, and 100B tokens. The $\times$ markers indicate the actual validation losses at a learning rate of $2\times10^{-3}$, which fall directly on the quadratic curves fitted using the remaining four learning rates.}
    \label{fig:moeproxy100bema}
    \vspace{-14pt}
\end{wrapfigure}

\subsubsection{Learning Rate Dynamics Across Token Horizons}\label{lrmup100B}
As Section \ref{lrmup} validates that $\mu$P enables learning rate transfer across width-scaled MoE models, we further investigate how the optimal learning rate shifts as the training token budget increases.
We define an MoE model with 5.6B total and 1.8B active parameters as the small-scale proxy model, and a $2\times$ width-scaled model with 20.7B total and 3.8B active parameters as the held-out target model.
Both models employ $\mu$P and are trained up to 100B tokens using a diverse data mixture that resembles realistic pretraining scenarios. 
Detailed model configurations are listed in Table \ref{tab:model-configs}.

Figure \ref{fig:moeproxy100bema} visualizes quadratic fits (Eq. \eqref{eq:parabola}) of validation loss against log-transformed learning rates ($6 \times 10^{-4}, \{1, 3, 6\} \times 10^{-3}$). 
As the token budget increases, the optimal learning rate exhibits a slight downward trend.
Notably, at each token scale, the fitted parabolas for both proxy and width-scaled target models exhibit highly consistent curvature and vertex locations, indicating that the optimal learning rate at a given token scale (i.e., the vertex of each parabola) transfers reliably across model widths.

\subsubsection{Predicting Optimal Learning Rate for Extremely Long Token Horizon}\label{lrmup10T}
In realistic large-scale MoE training scenarios, models are often pretrained over extremely long token horizons (e.g., over 10T tokens).
Accordingly, we extrapolate the findings from Section \ref{lrmup100B} to estimate the optimal learning rate for training at the 10T-token scale.

\begin{figure}[bt]
    \centering
    \begin{subfigure}[b]{0.49\textwidth}
        \centering
        \includegraphics[width=0.853\textwidth]{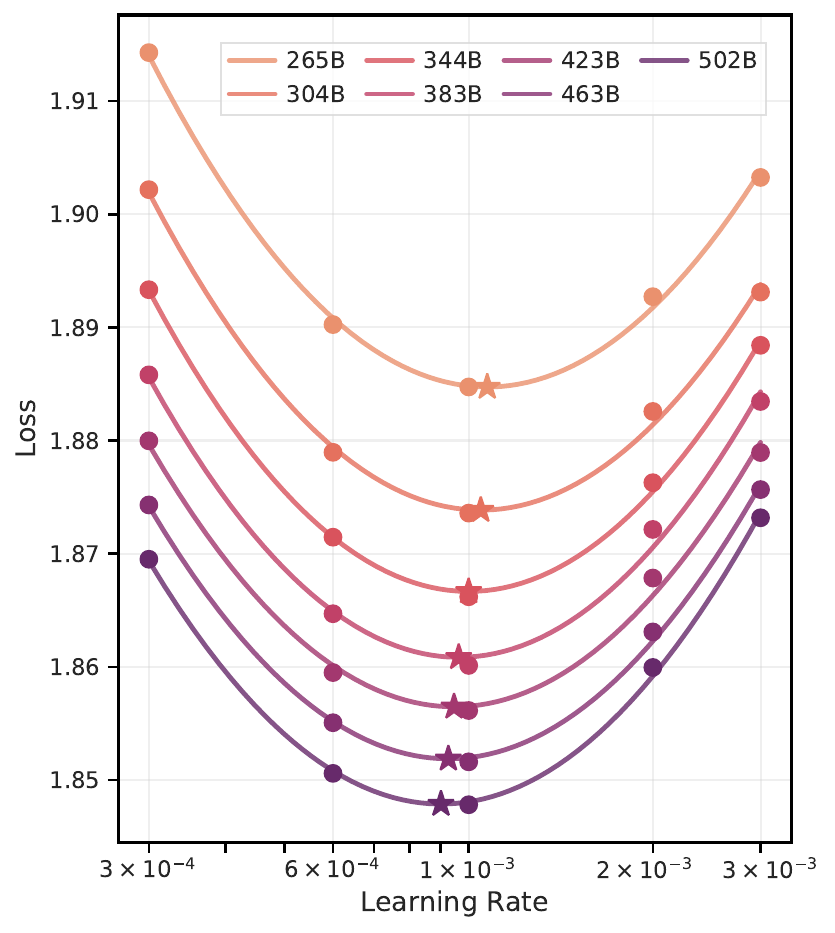} 
        \caption{Estimated Optimal LRs}
        \label{fig:moe500b-proxy-ema}
    \end{subfigure}
    \begin{subfigure}[b]{0.49\textwidth}
        \centering
        \includegraphics[width=\textwidth]{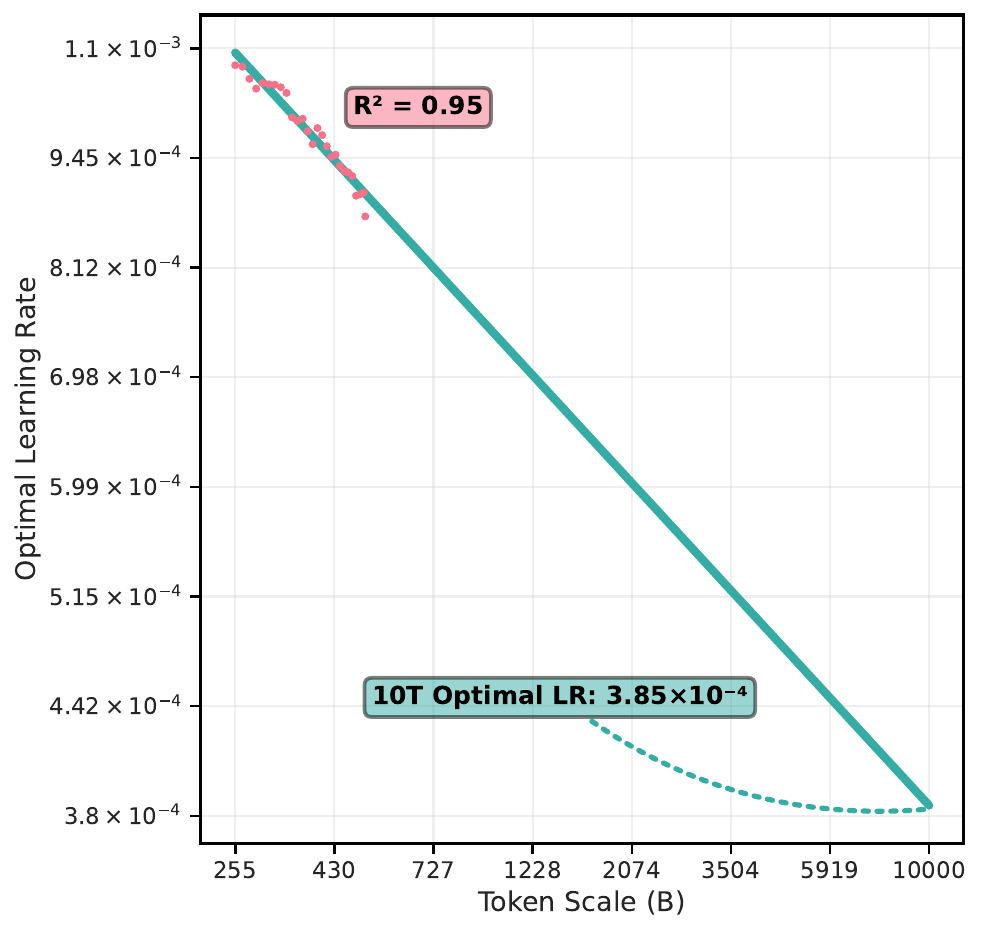} 
        \caption{Extrapolated Optimal LRs}
        \label{fig:moe10t-regression}
    \end{subfigure}
    
    \caption{(a) Second-order polynomials fitted between the log-transformed learning rate and validation loss for the base proxy MoE model (10.8B total, 3.3B active) at representative token scales (a subset of all scales is shown for visual clarity). The $\bigstar$ markers denote the optimal learning rates derived from the vertex of each parabola. (b) Linear regression in log-log space over these optimal learning rates across token scales ranging from 255B to 502B. The fitted model achieves $R^2=0.95$, enabling a reliable extrapolation of the optimal learning rate $3.85\times10^{-4}$ for the 10T-token pretraining.}
    \label{fig:moe-long-token-horizon}
    \vspace{-10pt}    
\end{figure}

\begin{wrapfigure}{rb}{0.5\textwidth}
    \vspace{-5pt}
    \centering
    \includegraphics[width=0.99\linewidth]{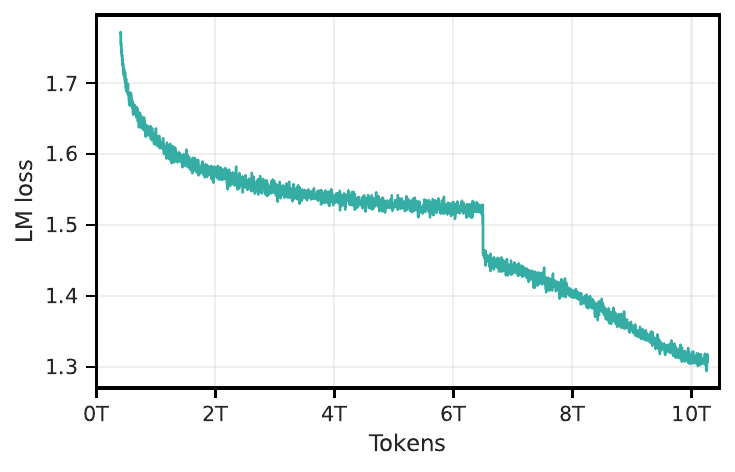}
    \caption{Training loss of our model (155B total, 17B active) during Stage 1 pretraining.}
    \label{fig:train-loss}
    \vspace{-15pt}
\end{wrapfigure}

We first train the base proxy model (10.8B total, 3.3B active; corresponding to $1/4\times$ width-scale of the target MoE model size, 155B total and 17B active parameters) for approximately 500B tokens. Both model configurations are listed in Table \ref{tab:model-configs} in Appendix \ref{Appendix:model-configs}. We then estimate the optimal learning rate at 10B-token intervals, utilizing EMA weights computed every 2B tokens.
Figure \ref{fig:moe500b-proxy-ema} illustrates the estimated optimal learning rates---derived from the vertices of the fitted parabolas---across varying token scales. 

Following the methodology in Section \ref{sec:2.2.2}, we fit a linear regression model to extrapolate the optimal learning rate for larger token horizons. Specifically, we restrict the fitted data points to token budgets after 255B, ensuring that training dynamics have stabilized following the batch size increase.
As shown in Figure \ref{fig:moe10t-regression}, the linear fit achieves a high $R^2$ score of 0.95, predicting an optimal learning rate of $3.85\times10^{-4}$ for 10T-token training. This demonstrates that the optimal learning rate can be reliably inferred by extrapolating the linear regression from small proxy runs. 
Additional empirical support for this extrapolation is provided by the held-out validation in Appendix \ref{Appendix:small-validation}.

\subsubsection{Applying to Our Foundation Model}\label{ourmodel}
To validate the effectiveness of our approach, we apply the proposed two-step hyperparameter transfer framework to predict the optimal learning rate for pretraining a large-scale MoE foundation model (155B total, 17B active parameters) over a 10T-token horizon. Detailed model configurations are provided in Table \ref{tab:model-configs}. 
As shown in Figure \ref{fig:main-2-b}, the full-scale pretraining of this MoE model requires approximately $98\times$ the total compute of the proxy runs used for optimal learning rate prediction.

\begin{figure}[t]
    \centering
    \includegraphics[width=\textwidth]{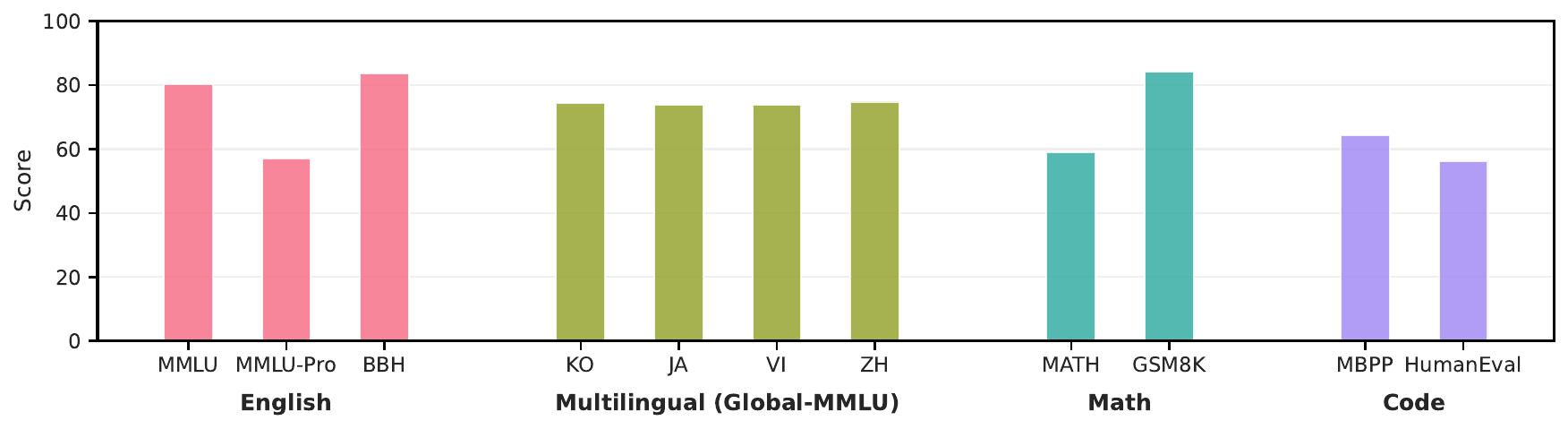} 
    \caption{Benchmark performance of our foundation MoE model (155B total, 17B active) after Stage 1 pretraining on 10T tokens. Scores are presented across four categorized domains: English (MMLU, MMLU-Pro, BBH), Multilingual (Global-MMLU: Ko, Ja, Vi, Zh), Math (MATH, GSM8K), and Code (MBPP, HumanEval).}
    \label{fig:performance}
    \vspace{-10pt}    
\end{figure}

\begin{wrapfigure}{rb}{0.5\textwidth}
    \vspace{-3pt}
    \centering
    \includegraphics[width=0.99\linewidth]{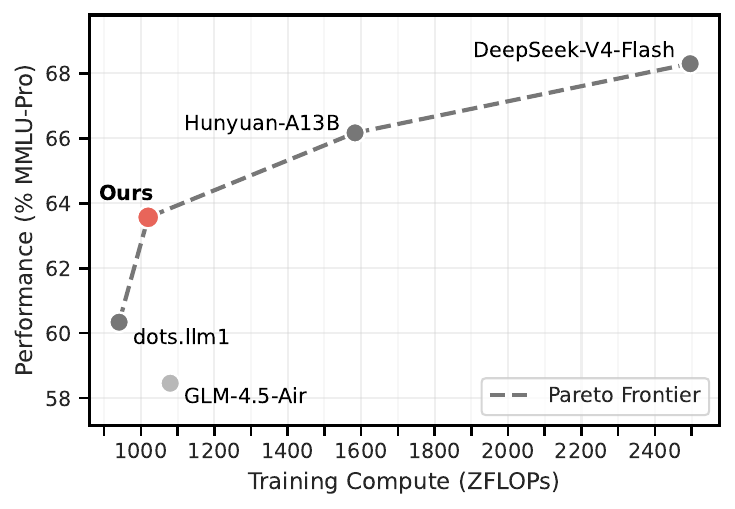}
    \caption{Estimated training compute vs. MMLU-Pro accuracy. Compute is estimated as $6ND$ with $N$ set to active parameters.}
    \label{fig:moe-pareto}
    \vspace{-15pt}
\end{wrapfigure}
The Stage 1 pretraining data mixture for our model initially consists of 45\% English, 12.5\% Math/STEM, 27.5\% Code, and 15.0\% Multilingual corpora. 
At the 6T-token mark, we adjust the mixture to 22.5\% English, 27.5\% Math/STEM, 25.0\% Code, and 25.0\% Multilingual corpora to enhance coverage of underrepresented domains.

Figure \ref{fig:train-loss} shows the training loss during Stage 1 pretraining, following batch size scheduling applied after 200B tokens. The pretraining loss of our foundation model remains highly stable throughout, with no loss spikes, confirming the validity of the extrapolated optimal learning rate. 
Evaluation results from Stage 1 pretraining are presented in Figure \ref{fig:performance}, with evaluation details available in Appendix \ref{Appendix:eval-details}. 
Given that the Stage 1 data mixture prioritizes domain diversity over strict quality filtering, these evaluation results confirm that the model was trained successfully and efficiently.

In addition, figure \ref{fig:moe-pareto} compares our model with publicly available open-weight MoE base models of comparable scale and active parameter count: dots.llm1 \citep{huo2025dots}, GLM-4.5-Air \citep{zeng2025glm}, Hunyuan-A13B \citep{hunyuan2025a13b}, and DeepSeek-V4-Flash \citep{xu2026deepseek}. 
Training compute is estimated using the standard $6ND$ approximation \citep{kaplan2020scaling,hoffmann2022training}, with $N$ set to active parameters and $D$ to the reported pretraining token count. 
For consistency, all models are evaluated using our evaluation harness. 
Figure \ref{fig:moe-pareto} shows that our model lies on the Pareto frontier, achieving higher MMLU-Pro \citep{wang2024mmlu} accuracy than dots.llm1 and GLM-4.5-Air at comparable or lower estimated training compute.






\section{Related Works}

\paragraph{Hyperparameter Transfer}
As the scale of foundation models expands, the computational cost of finding optimal hyperparameters becomes increasingly prohibitive. 
To mitigate this, recent studies such as Maximal Update Parameterization ($\mu$P) \citep{pmlr-v139-yang21c} and $\mu$-Transfer \citep{yang2022tensor} enable zero-shot transfer of optimal hyperparameters (e.g., learning rate) across model width by modifying parameterization schemes and scaling learning rates or logits for specific tensors. Building on this, \citet{wortsman2024small} and \citet{everett2024scaling} have further proposed and verified a simplified adaptation of $\mu$P.
While $\mu$P was initially applied to AdamW \citep{loshchilov2017decoupled}, it has recently been successfully extended to the Muon optimizer \citep{shah2025practical}.
However, existing studies that explore $\mu$P scaling for MoE models are mostly restricted to limited scenarios. 
For example, \citet{maasnicki2025muparametrization} attempt to apply $\mu$P to the Switch Transformer \citep{fedus2022switch} architecture, but their study was limited to the AdamW optimizer \citep{loshchilov2017decoupled} under restrictive settings, such as fixing the total and active expert counts and scaling only the hidden dimensions. 
These assumptions are difficult to generalize to large-scale MoE architectures that employ a large number of fine-grained experts (e.g., over 128), leaving reliable hyperparameter transfer for such powerful MoE models an open problem.
                                                 
\paragraph{Scaling Laws}
As model size and training data grow, exhaustive hyperparameter sweeps become increasingly impractical, motivating the study of scaling laws to predict the optimal configurations. 
For MoE models, existing scaling research has largely focused on identifying optimal architectural configurations as the parameter count scales \citep{clark2022unified,ludziejewski2024scaling}. 
Beyond model size, the training token horizon also significantly influences optimization dynamics, causing optimal hyperparameter configurations (e.g., learning rate, batch size) to shift. 
Recent studies have explored hyperparameter optimization based on computational budgets \citep{bi2024deepseek} and proposed extrapolating optimal configurations from small-scale proxies by jointly considering model and token scales \citep{bjorck2025scaling, li2025predictable}.
However, this joint scaling approach relies on sweeps across multiple dimensions (as depicted in Figures \ref{fig:main-a} and \ref{fig:main-2-a}). 

\section{Conclusion}
We propose a compute-efficient two-step framework for hyperparameter transfer in large-scale Mixture-of-Experts (MoE) pretraining, enabling accurate optimal learning rate estimation without exhaustive sweeps.
By combining $\mu$P-based width transferability with a linear scaling law across token budgets, we show that optimal learning rates for trillion-token large-scale MoE pretraining can be reliably predicted from small proxy experiments, thereby reducing unnecessary computation. 
We validate this methodology by pretraining our foundation model (155B total, 17B active parameters) from scratch. 
While conducting exhaustive full-scale sweeps to definitively verify the optimality of the predicted learning rate ($3.85\times10^{-4}$) is computationally infeasible, the exceptionally stable loss trajectory (Figure \ref{fig:train-loss}) and highly competitive benchmark scores (Figure \ref{fig:performance}) provide strong empirical evidence that our extrapolated configuration is effective and reliable for extreme-scale pretraining. 

In this study, we focus on MoE architectures with Multi-head Latent Attention (MLA) \citep{liu2024deepseekv2} and the Muon optimizer \citep{jordan2024muon}. 
Extending this framework to diverse MoE structures and other optimizers remains an important direction for future work.
Beyond architectural and optimizer generalization, we identify two additional research directions.
First, because top-$k$ routing can result in varying numbers of tokens assigned to individual experts, the effective batch size, gradient noise scale, and potentially the optimal update scale may vary across experts, suggesting that per-expert learning rate adaptation could yield further gains.
However, realizing this in practice requires accounting for how routing evolves during training, how data mixture recipes affect expert-level batches, and how such adaptation can be implemented efficiently at scale. 
Given the substantial experimental and engineering costs these considerations entail, we leave per-expert learning rate adaptation to future work.
Second, our large-scale recipe expands sparsity jointly with width axes rather than in isolation.
Although our results demonstrate successful hyperparameter transfer along this joint scaling path, the effect of the sparsity dimension itself cannot be clearly disentangled from that of width.
A controlled large-scale study of $\mu$P transfer along the sparsity axis alone would therefore be valuable. 
We anticipate that extending the core principles---width-based $\mu$P transfer and token-dimension scaling laws---to a broader range of settings will further reduce the computational burden of large-scale MoE research.




\bibliography{colm2026_conference}
\bibliographystyle{colm2026_conference}

\appendix

\section{Training Details}\label{Appendix:training-details}
We match Muon optimizer's update RMS to that of AdamW by scaling the learning rate by $0.2 \cdot \sqrt{\max (A,B)}$, where $A$ and $B$ are the dimensions of the full-rank matrix parameter (i.e., $\mathrm{fan\_in}$ and $\mathrm{fan\_out}$), following \citet{liu2025muon}. This adjustment allows for the direct transfer and reuse of the tuned learning rate and weight decay between the AdamW and Muon optimizers.
We use weight decay with a coefficient of $1\times10^{-1}$ and the auxiliary z-loss \citep{chowdhery2022palm} with a coefficient of $5\times10^{-6}$, both fixed as constants throughout all experiments. 
For validating $\mu$P for MoE with small-scale runs in Section \ref{lrmup}, we employ the cosine learning rate scheduler \citep{loshchilov2016sgdr} with a fixed minimum learning rate of $1\times10^{-5}$, varying only the maximum learning rate following \citet{wortsman2024small} and \citet{li2025predictable}. All other experiments use WSD scheduler \citep{hu2024minicpm}.

We use a fixed batch size of 0.5M tokens for experiments in Section \ref{lrmup} and 8M tokens for Section \ref{lrmup100B}.
For Section \ref{lrmup10T}, we increase the batch size from 8M tokens to 32M tokens after training on 200B tokens.

\begin{table*}[!t]
    \centering
    \resizebox{\textwidth}{!}{
    \begin{tabular}{c | c c c c c c c c }
    \toprule
    \textbf{Model Scale} & \textbf{Total Params.} & \textbf{Active Params.} & \textbf{$n_\text{layers}$} & \textbf{$d_\text{model}$} & \textbf{$d_\text{ff}$} & \textbf{$n_\text{heads}$} & \textbf{$n_\text{experts}$} & \textbf{$\lambda_\text{routed}$} \\ 
    \midrule
    \multicolumn{9}{c}{\textbf{Models for Section \ref{lrmup} and Appendix \ref{Appendix:LR-Dense}}} \\
    \midrule
    \textbf{Dense} Base proxy & 0.24B & 0.24B & 48 & 384 & 1536 & 4 & - & - \\ 
    \textbf{Dense} $2\times$ scale & 0.7B & 0.7B & 48 & 768 & 3072 & 8 & - & - \\     
    \textbf{Dense} $4\times$ scale & 2.27B & 2.27B & 48 & 1536 & 6144 & 16 & - & - \\     
    \textbf{Dense} $8\times$ scale & 8.02B & 8.02B & 48 & 3072 & 12288 & 32 & - & - \\  
    \midrule
    \textbf{MoE} Base proxy & 0.6B & 0.3B        & 48 & 256 & 6144 & 4 & 16 & 2.428 \\ 
    \textbf{MoE} $2\times$ scale & 2.2B & 0.7B  & 48 & 512 & 6144 & 8 & 32 & 2.442 \\ 
    \textbf{MoE} $4\times$ scale & 8B & 1.5B    & 48 & 1024 & 6144 & 16 & 64 & 2.446 \\ 
    \textbf{MoE} $8\times$ scale & 30.7B & 3.6B & 48 & 2048 & 6144 & 32 & 128 & 2.448 \\ 
    \midrule
    \multicolumn{9}{c}{\textbf{Models for Section \ref{lrmup100B}}} \\
    \midrule
    \textbf{MoE} Base proxy & 5.6B & 1.8B         & 32 & 1024 & 11264 & 16 & 32 & 2.442 \\ 
    \textbf{MoE} $2\times$ scale & 20.7B & 3.8B  & 32 & 2048 & 11264 & 32 & 64 & 2.446 \\ 
    \midrule
    \multicolumn{9}{c}{\textbf{Models for Figure \ref{fig:main-2}, Section \ref{lrmup10T} and Section \ref{ourmodel}}} \\
    \midrule
    \textbf{MoE} Base proxy & 10.8B & 3.3B      & 62 & 1024 & 12288 & 16 & 32  & 2.442 \\ 
    \textbf{MoE} $1.5\times$ scale & 23B & 4.9B  & 62 & 1536 & 12288 & 24 & 48 & 2.445 \\     
    \textbf{MoE} $2\times$ scale & 40.2B & 6.9B  & 62 & 2048 & 12288 & 32 & 64 & 2.446 \\     
    \textbf{MoE} $4\times$ scale & 155B & 17B  & 62 & 4096 & 12288 & 64 & 128 & 2.448 \\     
    \bottomrule
    \end{tabular}
    }
    \caption{Model configurations for scaling width. $n_\text{layers}$, $d_\text{model}$, $d_\text{ff}$, and $n_\text{heads}$ represent the number of layers, hidden dimension, intermediate dimension of an MLP (active experts for MoE; $d_\text{ff}=k\times d_\text{expert}$), and attention heads, respectively. $n_\text{experts}$ is the total number of experts, and $\lambda_\text{routed}$ denotes the MoE routed scaling factor.}
    \label{tab:model-configs}
\end{table*}

\section{Model Configuration Details}\label{Appendix:model-configs}
Table \ref{tab:model-configs} details the dense and MoE model configurations for both the base proxy model and width-scaled models. 
All models employ Multi-head Latent Attention (MLA) \citep{liu2024deepseekv2} for the attention mechanism. 
Model depth (i.e., the number of layers) is kept constant across all configurations. 
For the MoE models, we keep both the intermediate dimension of the expert MLP layers ($d_\text{expert}$) and the number of active experts ($k$) fixed during width-scaling. As a result, $d_\text{ff}=k\times d_\text{expert}$ remains unchanged.
Furthermore, for the experiments in Section \ref{lrmup10T} and Section \ref{ourmodel}, we replace the first layer of the MoE model with a single dense layer whose intermediate dimension equals the activated intermediate dimension of the MoE layers (i.e., $k\times d_\text{expert}$).

\section{Learning Rate Transfer with Dense Models}\label{Appendix:LR-Dense}
Since the combination of Multi-head Latent Attention (MLA) \citep{liu2024deepseekv2} and the Muon optimizer \citep{jordan2024muon} has not been extensively explored under the $\mu$P and learning rate scaling protocols, we first validate the efficacy of hyperparameter transfer on a dense model baseline before extending the analysis to Mixture-of-Experts (MoE) architectures. 
For MLP layers in dense models, all fully connected layers are considered as the matrix-like hidden weights (Table \ref{tab:params}).

\begin{figure}[t]
    \centering
    \begin{subfigure}[b]{0.49\textwidth}
        \centering
        \includegraphics[width=\textwidth]{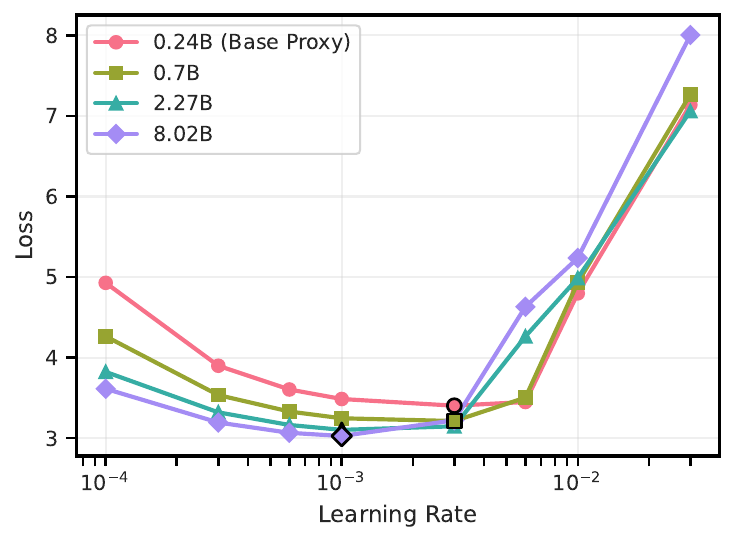} 
        \caption{Standard Parameterization (SP)}
        \label{fig:350m}
    \end{subfigure}
    \begin{subfigure}[b]{0.49\textwidth}
        \centering
        \includegraphics[width=\textwidth]{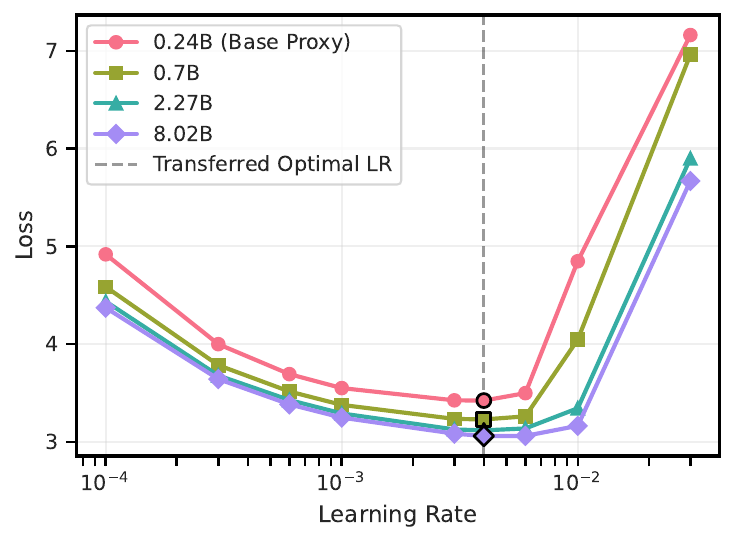} 
        \caption{Maximal Update Parameterization ($\mu$P)}
        \label{fig:1.7b}
    \end{subfigure}
    
    \caption{Comparison of learning rate transferability under Standard Parameterization (SP) and Maximal Update Parameterization ($\mu$P) across dense models of increasing width. Markers with black outlines indicate the learning rate achieving the lowest training loss (i.e., optimal learning rate) for each model. (a) Under SP, the optimal learning rate diverges across varying model widths. (b) Under $\mu$P, the optimal learning rate identified in the base proxy (0.24B) consistently transfers across models scaled to $2\times$ (0.7B), $4\times$ (2.27B), and $8\times$ (8.02B) the base width, demonstrating robust zero-shot hyperparameter transfer in dense MLA models as well as MoE models.}
    \label{fig:densemup}
\end{figure}

We configure the base proxy model with a hidden dimension ($d_\text{model}$) of 384, an intermediate dimension ($d_\text{ff}$) of 1536 (four times the hidden dimension), and 4 attention heads. 
We then scale the model width to $d_\text{model}\in\{768, 1536, 3072\}$, $d_\text{ff}$ to $\{3072, 6144, 12288\}$, and the number of attention heads to $\{8, 16, 32\}$, corresponding to parameter counts of 0.24B (base proxy), 0.7B, 2.27B, and 8.02B.
All models have 48 layers and are trained for 1.3B tokens.
For each model size, we sweep the maximum learning rate over $\{1, 3, 6\} \times 10^{-4}, \{1, 3, 6\} \times 10^{-3},$ and $\{1, 3\} \times 10^{-2}$ to validate transferability.
Figure \ref{fig:densemup} demonstrates that under $\mu$P, the optimal learning rate that minimizes training loss is preserved across width-scaled Dense MLA models.

\section{Evaluation Details}\label{Appendix:eval-details}
For evaluating our model after Stage 1 pretraining (10T tokens), we use a customized evaluation framework based on \citet{eval-harness}. Benchmarks are categorized as follows:

\paragraph{English General Knowledge:} MMLU (5-shots) \citep{hendrycks2020measuring}, MMLU-Pro (5-shots) \citep{wang2024mmlu}, BBH (3-shots) \citep{suzgun2023challenging}

\paragraph{Multilingual General Knowledge:} Global-MMLU (Korean, Japanese, Vietnamese, Chinese; 5-shots) \citep{singh2025global}

\paragraph{Math:} MATH (4-shots) \citep{lewkowycz2022solving}, GSM8K (8-shots) \citep{cobbe2021training}

\paragraph{Code:} MBPP (Pass$@$1, 3-shots) \citep{austin2021program}, HumanEval (Pass$@$1, 0-shot) \citep{chen2021evaluating}

\section{Small-Scale Validation of Learning Rate Extrapolation}\label{Appendix:small-validation}
Our two-step framework (Section \ref{sec:2}) predicts the optimal learning rate (LR) at a long token horizon from a short-budget LR sweep. 
A direct validation at the 10T-token target is computationally infeasible: even for a proxy model at one-tenth the size of the target model (155B total, 17B active parameters), sweeping five learning rates for 10T tokens would consume roughly half the compute of the full 155B model training.
Therefore, we perform a retrospective held-out validation using the LR sweep data from Section \ref{lrmup10T}.

Specifically, among the token-budget data points in the 255B--500B range, we use only the first 11 budgets (approximately 255B--350B) to fit the extrapolation model. 
Following the same procedure as in the main paper, we first estimate the optimal learning rate at each budget using Eq. \eqref{eq:parabola}, and then fit the log-linear regression in Eq. \eqref{eq:extrapolate}. 
The resulting model is used to predict the optimal learning rates for five unseen token budgets near 500B, which are then compared against the optimal learning rates obtained by independently fitting Eq. \eqref{eq:parabola} to the actual LR and loss values at each held-out budget.

\begin{table}[h]
\centering
\small
\begin{tabular}{lccc}
\toprule
Token Budget & Predicted Optimal LR & Actual Optimal LR & Ratio \\
\midrule
462.7B & $9.57\times10^{-4}$ & $9.26\times10^{-4}$ & $1.033\times$ \\
472.6B & $9.53\times10^{-4}$ & $9.22\times10^{-4}$ & $1.034\times$ \\
482.5B & $9.49\times10^{-4}$ & $8.97\times10^{-4}$ & $1.058\times$ \\
492.4B & $9.45\times10^{-4}$ & $8.99\times10^{-4}$ & $1.052\times$ \\
502.3B & $9.41\times10^{-4}$ & $9.01\times10^{-4}$ & $1.045\times$ \\
\bottomrule
\end{tabular}
\caption{Held-out validation of learning rate extrapolation. The extrapolation model is fitted using only token budgets in 255B--350B and evaluated on unseen budgets near 500B. Ratios refer to Predicted Optimal LR divided by Actual Optimal LR.}
\label{tab:heldout_lr}
\end{table}

As shown in Table \ref{tab:heldout_lr}, the predicted optimal learning rates (Predicted Optimal LR) closely match the independently estimated optima using Eq. \eqref{eq:parabola} (Actual Optimal LR), with an average discrepancy of $\approx 4.4\%$.
The gap is substantially smaller than the extrapolation errors reported by \citet{bjorck2025scaling}. 
These results provide evidence that the proposed token-budget extrapolation accurately identifies the near-optimal learning-rate region over the range relevant to our two-step scaling procedure.

\section{Additional Analysis of Expert Routing Bias in Staged Pretraining}\label{Appendix:staged-routing}
Having successfully completed Stage 1 pretraining of our foundation model (Section \ref{ourmodel}), we subsequently conducted Stage 2 training on a smaller-scale, high-quality dataset. During this process, we observed an interesting phenomenon related to expert routing dynamics that we share as an additional finding.

Staged pretraining—typically consisting of a large-scale general-domain training (Stage 1) followed by smaller-scale, high-quality, and domain-augmented training (Stage 2)—has been shown to be effective for training large language models from scratch \citep{hu2024minicpm,blakeney2024does,abdin2024phi,walsh20252,allal2025smollm2}.
However, the resulting shift in data distribution may introduce instability in expert routing for MoE models.

\begin{figure}[tb]
    \centering
    \begin{subfigure}[b]{0.49\textwidth}
        \centering
        \includegraphics[width=\textwidth]{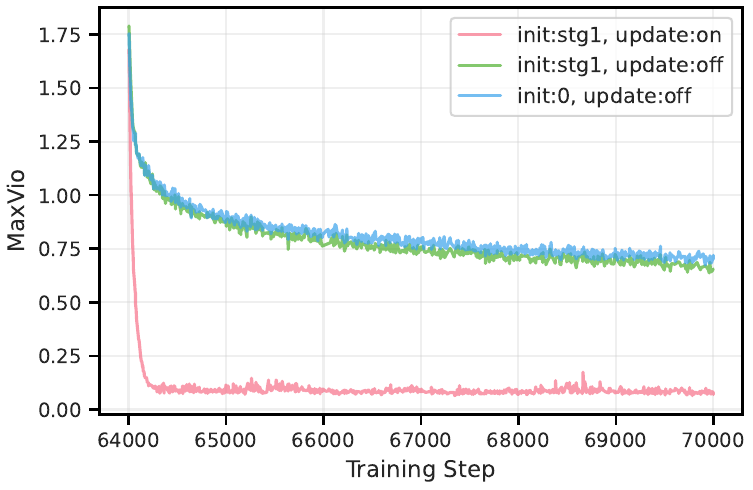} 
        \caption{Average MaxVio}
        \label{fig:avg_maxvio}
    \end{subfigure}
    \hfill 
    \begin{subfigure}[b]{0.49\textwidth}
        \centering
        \includegraphics[width=\textwidth]{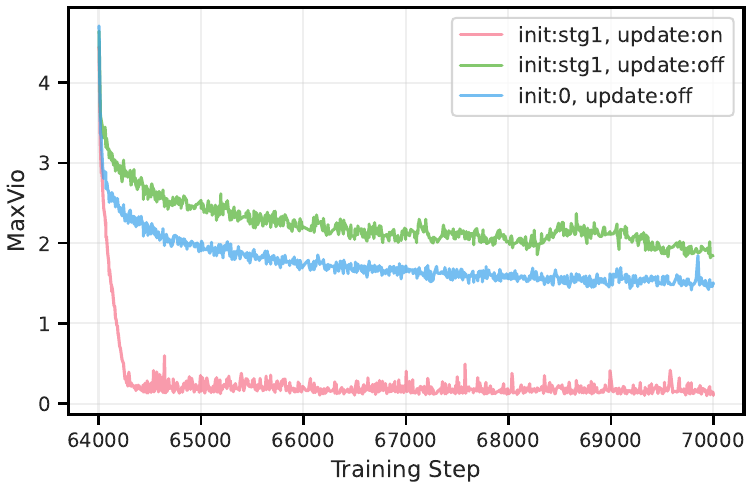} 
        \caption{Max MaxVio}
        \label{fig:max_maxvio}
    \end{subfigure} 
    
    \caption{MaxVio trends for three different expert routing bias settings during Stage 2 training. (a) Average MaxVio across MoE layers. (b) Maximum MaxVio among all MoE layers at each training step.}
    \label{fig:stg2_bias_abl_avg_max_maxvio}
\end{figure}
As we adopt the auxiliary-loss-free load balancing strategy \citep{liu2024deepseek}, expert routing bias and the sequence-level load balancing loss are the primary factors stabilizing expert routing dynamics.
Since sequence-level load balancing loss operates independently on each sequence irrespective of the underlying data distribution, we focus exclusively on factors related to expert bias. 
Specifically, we consider three expert bias configurations during Stage 2 pretraining:
(1) initializing the expert bias with values obtained at the end of Stage 1 and continuing updates during Stage 2;
(2) initializing with Stage 1 values but freezing updates during Stage 2;
and (3) removing the expert bias terms entirely (initializing to zero and disabling updates).

Figure \ref{fig:stg2_bias_abl_avg_max_maxvio} shows notable differences in MaxVio \citep{wang2024auxiliary}, a metric that reflects expert utilization imbalance:
\begin{equation}\label{eq:maxvio}
\mathrm{MaxVio} = \frac{\max_i\mathrm{Load}_i-\overline{\mathrm{Load}_i}}{\overline{\mathrm{Load}_i}}
\end{equation}
where $\mathrm{Load}_i$ denotes the number of tokens routed to the $i$-th expert, and $\overline{\mathrm{Load}_i}$ stands for the expected number of tokens routed under perfect load balancing. 
MaxVio is computed at the global-batch level, and we report both the average and maximum values across MoE layers.
\begin{wrapfigure}{r}{0.5\textwidth}
    \vspace{5pt}
    \centering
    \includegraphics[width=0.98\linewidth]{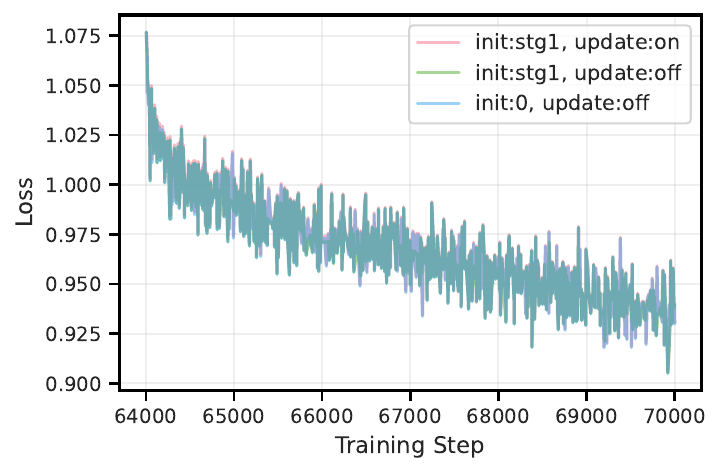}
    \caption{All three expert routing bias settings exhibit nearly identical training loss trends.}
    \label{fig:stg2_bias_abl_loss}
    \vspace{-15pt}
\end{wrapfigure}

Our results indicate that continuously updating the expert bias during Stage 2 leads to the most balanced expert utilization. 
Interestingly, when expert bias updates are frozen, zero initialization is more beneficial for balanced expert utilization than inheriting values from Stage 1.
This suggests that expert bias terms optimized for Stage 1 data distribution may not be suitable for the shifted distribution in Stage 2.
Nevertheless, Figure \ref{fig:stg2_bias_abl_loss} shows nearly identical training loss trajectories across all three configurations. This indicates that expert bias terms have a negligible impact on overall model optimization, implying that severe expert routing collapse is unlikely during Stage 2 regardless of the bias configurations.

\section{Layer-wise Routing Balance and Expert Specialization}\label{Appendix:expert-specialization}
To further examine whether the relatively small MaxVio values of our foundation model (Section \ref{ourmodel}) during Stage 2 training (Figure \ref{fig:stg2_bias_abl_avg_max_maxvio}) indicate overly balanced routing, we perform a layer-wise analysis of routing behavior.
We evaluate routing on 200 prompts per benchmark (except for HumanEval, where all 164 available problems are used), drawn from four domain families: \emph{English} (MMLU \citep{hendrycks2020measuring}, MMLU-Pro \citep{wang2024mmlu}, BBH \citep{suzgun2023challenging}), \emph{Multilingual} (Global-MMLU \citep{singh2025global}: Ko, Ja, Vi, Zh), \emph{Math} (MATH \citep{hendrycks2021measuring}, GSM8K \citep{cobbe2021training}), and \emph{Code} (MBPP \citep{austin2021program}, HumanEval \citep{chen2021evaluating}).
For the \emph{Code} domain, each prompt is concatenated with its reference solution so that routing is measured on tokens that include actual code.
Benchmarks within the English, Math, and Code families are pooled together for measuring expert specialization, whereas each Global-MMLU language is treated separately.

We report three complementary metrics measuring marginal load imbalance and conditional expert specialization.
They are defined as follows.

\paragraph{(i) Aggregate MaxVio.}
We use the same MaxVio definition as in Appendix \ref{Appendix:staged-routing} (Eq. \eqref{eq:maxvio}), computed on the whole evaluation set pooled across all target benchmarks. 
We refer to this as the Aggregate MaxVio. 
Lower Aggregate MaxVio indicates more balanced routing across experts.

\paragraph{(ii) Normalized mutual information $I(E;D)/H(D)$.}
Let $\mathcal{D}$ denote the set of domains, and let $E$ and $D$ denote the random variables for the selected expert and the domain label, respectively.
For each domain $d \in \mathcal{D}$, let $p(e\mid d)$ be the conditional routing distribution over experts.
We adopt a uniform domain prior $p(d) = 1/|\mathcal{D}|$, independent of per-domain token counts, so that larger domains do not dominate the statistic. 
Under this prior, the domain entropy is $H(D)=\log|\mathcal{D}|$, and the marginal routing distribution is $\bar{p}(e)=\tfrac{1}{|\mathcal{D}|}\sum_{d \in \mathcal{D}} p(e\mid d)$. 
The mutual information between expert selection and the domain label is
\begin{equation}
I(E;D) = \frac{1}{|\mathcal{D}|}\sum_{d \in \mathcal{D}} 
D_\mathrm{KL}\!\left(p(e\mid d)\,\|\,\bar{p}(e)\right)
\end{equation}
We normalize it by $H(D)$,
\begin{equation}
I/H = \frac{I(E;D)}{H(D)} = \frac{I(E;D)}{\log|\mathcal{D}|} \in [0,1]
\end{equation}
$I/H = 0$ means every domain uses experts with the same distribution, whereas $I/H = 1$ means expert selection perfectly identifies the domain. 
Intermediate values quantify the fraction of domain entropy resolved by expert routing—e.g., $I/H = 0.15$ indicates that routing resolves about 15\% of the domain entropy and should not be interpreted as domain-classification accuracy.
Higher values indicate stronger domain-dependent expert specialization.

\paragraph{(iii) Mean pairwise Jensen--Shannon divergence.}
For any two domains $a$ and $b$ with routing distributions $p(e\mid a)$ and $p(e\mid b)$, let $m = \tfrac{1}{2}\bigl(p(e\mid a) + p(e\mid b)\bigr)$. Then
\begin{equation}
\mathrm{JSD}(a,b) 
= \tfrac{1}{2}D_\mathrm{KL}\!\left(p(e\mid a)\,\|\,m\right)
+ \tfrac{1}{2}D_\mathrm{KL}\!\left(p(e\mid b)\,\|\,m\right)
\end{equation}
The reported mean pairwise JSD is the average of $\mathrm{JSD}(a,b)$ over all domain pairs.
Using natural logarithms, the mean pairwise JSD is bounded above by $\log 2 \approx 0.693$. 
A mean pairwise JSD value of $0$ means every pair of domains has identical expert routing distributions.
Higher values indicate more distinct expert routing patterns across different domains.

\begin{figure}[t]
    \centering
    \includegraphics[width=\textwidth]{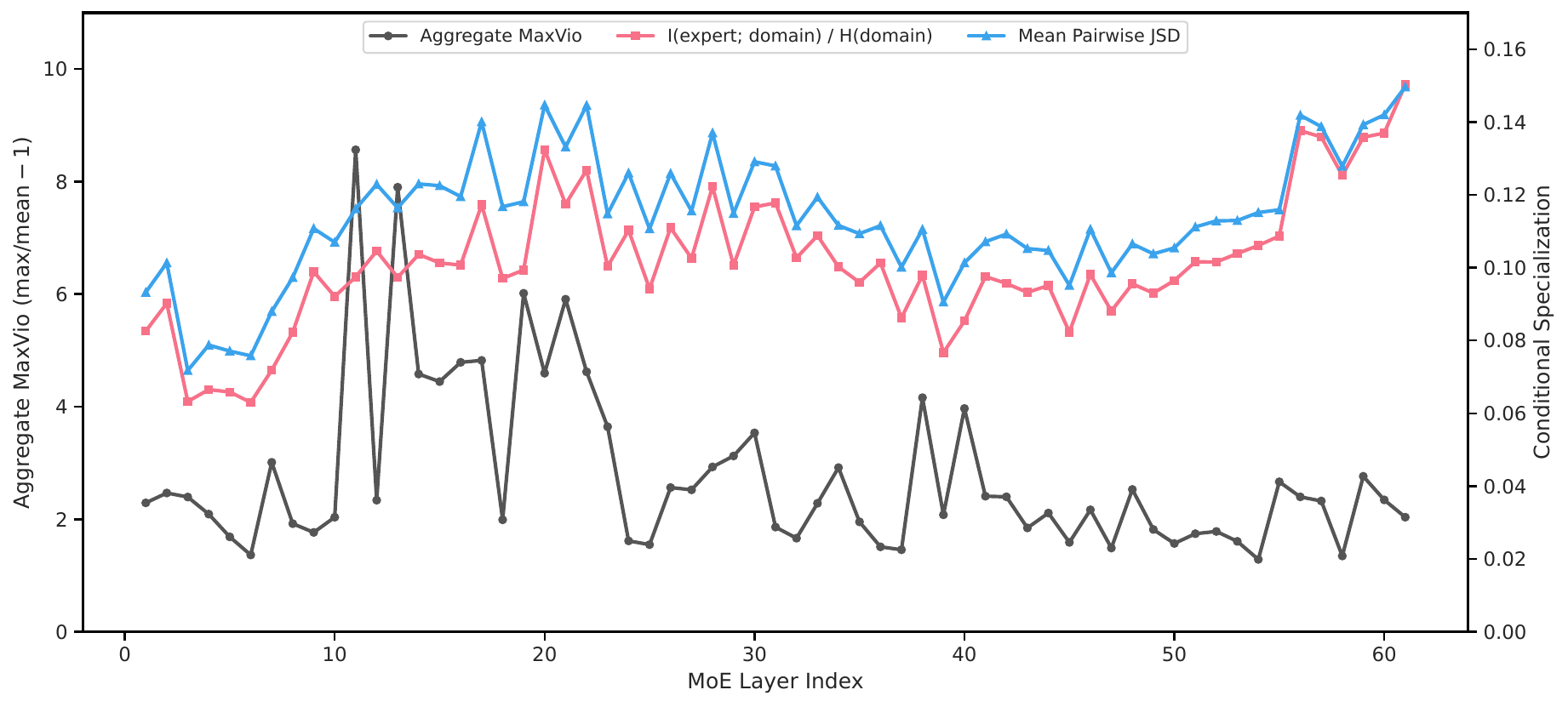} 
    \caption{Aggregate MaxVio (left axis) measures marginal expert-load imbalance, while normalized mutual information ($I/H$) and mean pairwise Jensen–Shannon divergence (right axis) measure domain-dependent routing specialization. Specialization increases toward deeper MoE layers, whereas MaxVio follows a different trend, demonstrating that balanced marginal routing does not necessarily imply suppressed specialization.}
    \label{fig:exp_specialized}
\end{figure}
Figure \ref{fig:exp_specialized} shows that marginal expert load balance and conditional specialization behave differently across depth.
Aggregate MaxVio remains relatively stable throughout most MoE layers, with only a localized increase in the shallow-to-middle layers.
In contrast, both conditional specialization measures ($I/H$, mean pairwise JSD) increase toward deeper layers and reach their highest values in the final MoE layers. 
This demonstrates that balanced marginal routing does not imply weak expert specialization: experts can maintain balanced overall utilization while simultaneously exhibiting domain-dependent routing preferences.
Note that the Aggregate MaxVio values here are computed on downstream evaluation benchmarks and are therefore not directly comparable to the training-time MaxVio in Figure \ref{fig:stg2_bias_abl_avg_max_maxvio}; these analyses should be interpreted as complementary evidence of layer-wise routing behavior and expert specialization, rather than a direct numerical comparison.

\begin{figure}[t]
    \centering
    \includegraphics[width=\textwidth]{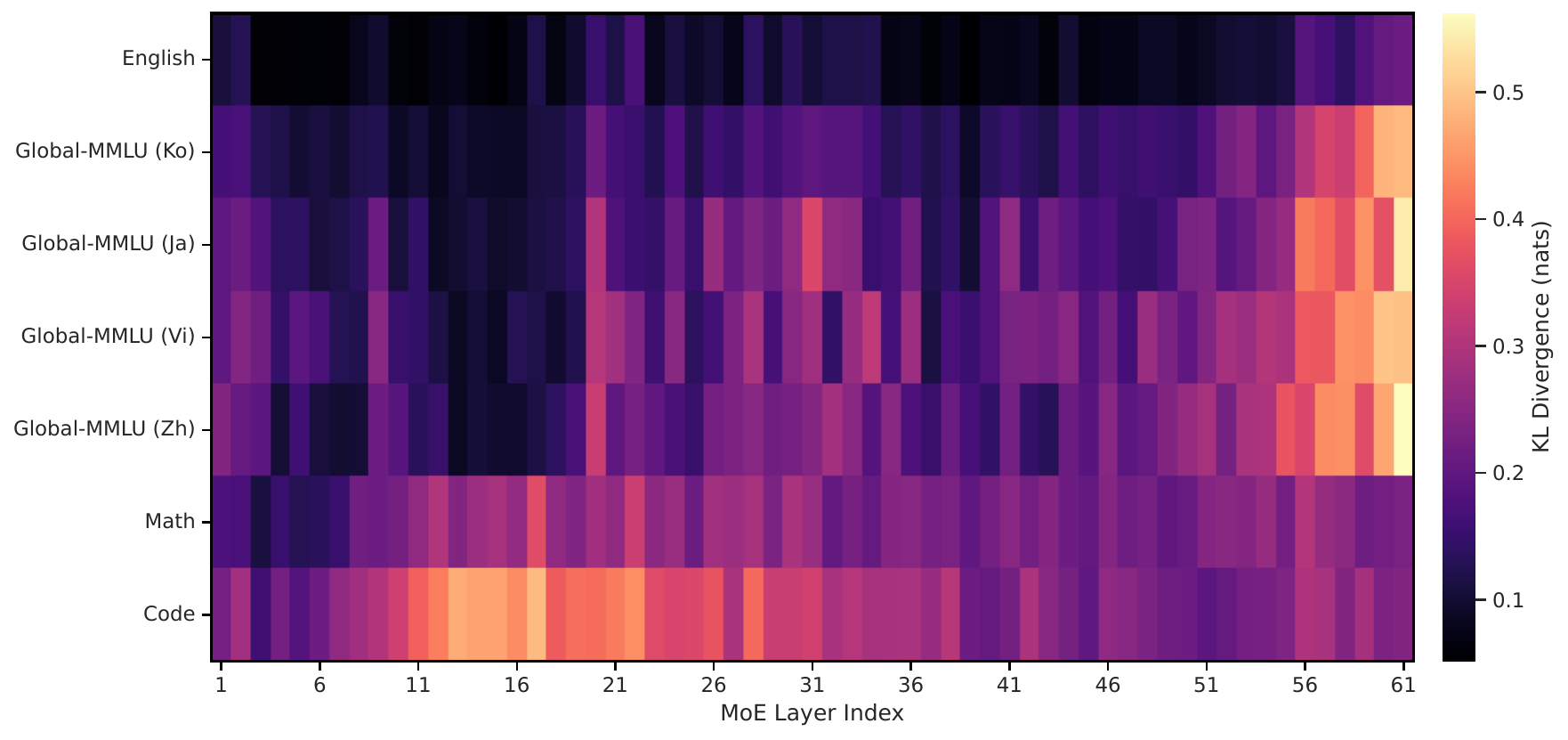} 
    \caption{Domain-conditioned expert routing divergence across MoE layers. Each cell shows  $D_{\mathrm{KL}}\!\left(p(e\mid d)\,\|\,\bar{p}(e)\right)$, where $\bar{p}(e)$ is the layer-wise marginal routing distribution defined in (ii). Brighter colors indicate stronger domain-specific routing.}
    \label{fig:exp_kl}
\end{figure}

To further illustrate the depth dependence of specialization, Figure \ref{fig:exp_kl} visualizes the KL divergence $D_{\mathrm{KL}}\!\left(p(e\mid d)\,\|\,\bar{p}(e)\right)$, where $p(e\mid d)$ denotes the expert routing distribution conditioned on domain $d$ and $\bar{p}(e)$ is the marginal routing distribution aggregated across domains at the same layer. 
Larger values indicate stronger domain-specific routing relative to the shared marginal distribution.

The heatmap (Figure \ref{fig:exp_kl}) reveals distinct specialization patterns across domains. 
Code routing deviates from the marginal distribution in the early layers, whereas multilingual routing remains close to the marginal until the final MoE layers, where specialization increases sharply. 
Together with Figure \ref{fig:exp_specialized}, these results show that balanced marginal routing can coexist with domain-dependent expert specialization.

\end{document}